\copyrightyear{2025}
\copyrightclause{Copyright for this paper by its authors.
  Use permitted under Creative Commons License Attribution 4.0
  International (CC BY 4.0).}

\conference{MediaEval'25: Multimedia Evaluation Workshop,
  October 25--26, 2025, Dublin, Ireland and Online}

\title{Multi-Task Learning for Visually Grounded Reasoning in Gastrointestinal VQA} 

\author[1]{Itbaan Safwan}[%
  email=i.safwan.26197@khi.iba.edu.pk,
]
\author[1]{Muhammad Annas Shaikh}[%
  email=m.shaikh.26919@khi.iba.edu.pk,
]
\author[1]{Muhammad Haaris}[%
  email=m.haaris.27083@khi.iba.edu.pk,
]

\author[1]{Ramail Khan}[%
  email=r.khan.26924@khi.iba.edu.pk,
]

\author[1]{Muhammad Atif Tahir}[%
  email=atiftahir@iba.edu.pk,
]

\address[1]{School of Mathematics and Computer Science, Institute of Business Administration (IBA), Karachi, Pakistan}

\begin{abstract}
We present a multi-task framework for the MediaEval Medico 2025 challenge, leveraging a LoRA-tuned Florence-2 model for simultaneous visual question answering (VQA), explanation generation, and visual grounding. The proposed system integrates three curated datasets: (1) Kvasir-VQA-x1 for question-answer learning, (2) a synthetically enriched explanation dataset offering structured medical reasoning, and (3) text-to-region pairs linking visual features with segmentation masks. This multi-task setup enables the model to jointly learn visual grounding, reasoning, and interpretation, producing responses that are both accurate and interpretable. Extensive evaluation demonstrates that our approach substantially improves over single-task baselines in both answer accuracy and visual localization, highlighting the effectiveness of grounded multi-task learning for medical VQA applications. Our code: \href{https://github.com/itbaans/multi_task_training_gastrointestinal_vqa}{\texttt{GitHub Repository}}

\end{abstract}

\maketitle

\section{Introduction}
The MediaEval Medico 2025 VQA Challenge~\cite{Medico2025} focuses on explainable AI for Gastrointestinal imaging. It features two subtasks—visual question answering and multimodal explanation generation—based on the Kvasir-VQA-x1 dataset~\cite{KvasirVQAx1}.

Building on this challenge, we address a key limitation of Vision-Language Models (VLMs): while VLMs recognize visual features well, they often lack medical terminology and reasoning for trustworthy clinical applications. Standard VQA training without explicit visual grounding or detailed explanations is insufficient for medical AI. We propose a multi-task learning framework enriching training with textual explanations and visual grounding through two curated datasets: (1) textual explanations providing medical reasoning for complexity-1 questions, and (2) visually grounded data linking answers to segmentation masks. Training Florence-2~\cite{xu2024florence} with this framework enables simultaneous learning of visual grounding, medical reasoning, and question answering, improving VQA performance while producing models grounded in visual evidence and medical knowledge.

The principles guiding our framework draw from prior work that has consistently shown multi-task learning to be effective for vision-language models. \cite{lu202012in1} trained on 12 tasks, reducing parameters from 3B to 270M while improving performance by 2.05 points. Molmo~\cite{deitke2024molmopixmoopenweights} showed that curating detailed captioning, QA pairs, and 2D pointing enables smaller models to outperform larger closed-source models. In medical imaging,~\cite{rui2025multimodalvisionpretrainingmedical} achieved Dice score gains of 0.28\%-14.47\% using multi-modal pre-training on 2.4M brain MRI images. We apply these principles to medical VQA, combining visual grounding, explanation generation, and question
 answering for GI image analysis.

\begin{figure}[!ht]
    \centering
    \includegraphics[width=0.8\textwidth]{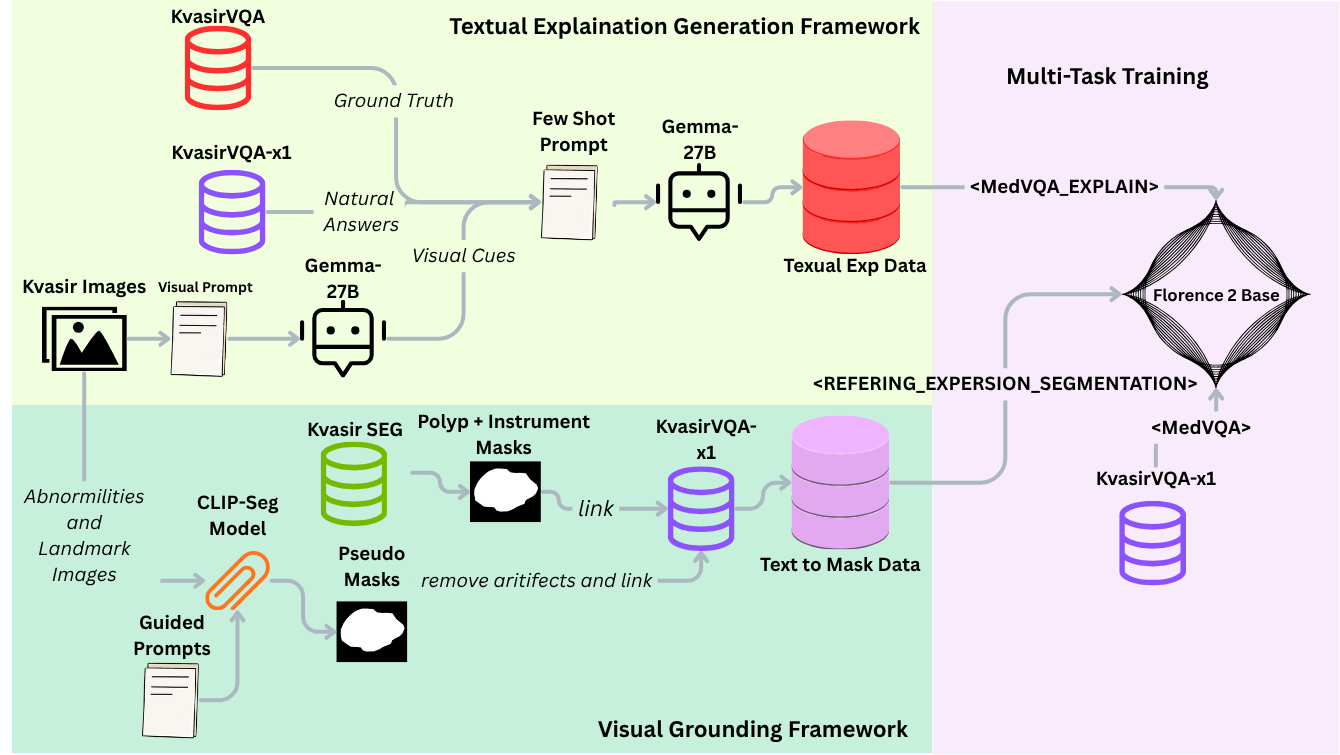}
    \caption{Our proposed multi-task training framework.}
    \label{fig:framework}
\end{figure}

\section{Methodology}

\subsection{Textual and Visual Generation Framework}
\textbf{Textual Explanation Dataset} was curated in two stages. First, starting with simple image-QA pairs from \cite{KvasirVQAx1}, we enriched them with ground-truth metadata (e.g., abnormality type, location) from \cite{KvasirVQA}. We then used the Gemma-27B API~\cite{gemma2024} to generate further descriptive visual cues. The prompting strategy for Gemma explicitly discouraged medical terminology, focusing instead on observable features (shape, texture) to improve generalization. In the second stage, we used the Gemma-27B model with few-shot prompting~\cite{brown2020languagemodelsfewshotlearners} to synthesize the ground-truth metadata, visual descriptions, and original QA pairs into complete, well-structured textual explanations. After post-processing to enhance fluency, the final dataset contained approximately 3,344 samples.

\textbf{Text-to-Relevant-Region Dataset} was curated by first collecting segmentation masks. We generated pseudo-masks for abnormalities and landmarks (2,954 samples) using the ClipSeg model \cite{lüddecke2022imagesegmentationusingtext} with descriptive text prompts (e.g., ``red patches''). To improve coverage, we used multiple prompts per case and refined the resulting masks by removing black backgrounds and artifacts using OpenCV. We also incorporated 1,383 high-quality polyp and instrument masks from the Kvasir-SEG datasets~\cite{jha2020kvasirseg, 10.1007/978-3-030-67835-7_19}. In the second stage, we linked these masks to corresponding textual answers from \cite{KvasirVQAx1}. The final dataset comprises 4,337 text–mask pairs. Although pseudo-masks lack medical precision, they offer useful approximate supervision, guiding the model to ground predictions in relevant visual evidence.

\subsection{Training and Post-processing}
The Florence-2 model was fine-tuned using LoRA on a combined dataset (Kvasir-VQA-x1, textual explanation, text-to-region) with an 80/20 split to prevent image overlap between training and validation sets. Training was conducted on Kaggle (2$\times$T4 GPUs) for one epoch using the Transformers Trainer library. We applied LoRA (rank=128, alpha=256) to all attention modules and trained with a learning rate of $5\times10^{-5}$, warmup\_ratio=0.1, FP16 precision, and an effective batch size of 12 (2 per device × 3 accumulation steps).

For post-processing and inference, we used three task-specific tokens: \texttt{<MedVQA>} for VQA, \texttt{<MedVQA\_EXPLAIN>} for explanations, and \texttt{<REFERRING\_EXPRESSION\_SEGMENTATION>} for region grounding (masks were converted to Florence-2 location tokens). \textbf{Subtask~1} generated outputs using \texttt{<MedVQA>} with the question as input. \textbf{Subtask~2} followed a two-step approach: first predicting the answer with \texttt{<MedVQA>}, then localizing the region using \texttt{<REFERRING\_EXPRESSION\_SEGMENTATION>} with the predicted answer. Explanations were generated using \texttt{<MedVQA\_EXPLAIN>} with the question and the prompt ``Explain in detail''. A confidence score for each generated explanation was derived from decoding stability, computed as the average top-$k$ ($k=5$) probability mass over top tokens. We found that conflicting explanations scored lower, while correct ones scored higher, though further validation is needed.

\begin{table}[!htbp] \centering
\caption{Validation Scores Across Different Model Variants. FL2: Florence-2; MT: Multi-task Training; Format: Rank\_Alpha (e.g., 32\_64 indicates LoRA rank=32, alpha=64)}
\label{tab:validation_scores}
\resizebox{\textwidth}{!}{%
\begin{tabular}{@{}lccccccccc@{}}
\toprule
\textbf{Model Configuration} & \textbf{Seg-IoU} & \textbf{Seg-IoU} & \textbf{Seg-IoU} & \textbf{BLEU} & \textbf{ROUGE-1} & \textbf{ROUGE-2} & \textbf{ROUGE-L} & \textbf{METEOR} \\
& \textbf{Instrument} & \textbf{Polyp} & \textbf{Pseudo} & & & & & \\
\midrule
FL2\_VQA\_32\_64 & 0.4911 & 0.4029 & 0.1309 & 0.4348 & 0.6856 & 0.4935 & 0.6566 & 0.6548 \\
FL2\_VQA\_64\_128 & 0.3712 & 0.5589 & 0.1651 & 0.4518 & 0.6953 & 0.5074 & 0.6675 & 0.6674 \\
FL2\_VQA\_128\_256 & 0.2961 & 0.2390 & 0.1688 & 0.4623 & 0.7024 & 0.5163 & 0.6754 & 0.6735 \\
\midrule
FL2\_VQA\_MT\_32\_64 & 0.7098 & 0.6828 & 0.5110 & 0.4432 & 0.6894 & 0.5010 & 0.6613 & 0.6592 \\
FL2\_VQA\_MT\_64\_128 & 0.7374 & \textbf{0.7063} & 0.5344 & 0.4615 & 0.7008 & 0.5144 & 0.6731 & 0.6728 \\
FL2\_VQA\_MT\_128\_256 & \textbf{0.7403} & 0.6879 & \textbf{0.5447} & \textbf{0.4726} & \textbf{0.7076} & \textbf{0.5234} & \textbf{0.6802} & \textbf{0.6804} \\
\midrule
\multicolumn{9}{l}{\textbf{\textit{Official results on Private Dataset (FL2\_VQA\_MT\_128\_256):}}} \\
FL2\_VQA\_MT\_128\_256\tnote{a} & -- & -- & -- & 0.4539 & 0.6828 & 0.4954 & 0.6531 & 0.6515 \\
\bottomrule
\end{tabular}%
}
\begin{tablenotes}
    \small 
    \item[a] Additional results on private dataset: CHRF++ = 63.79, BERTScore F1 = 0.9479.
\end{tablenotes}
\end{table}

\begin{figure}[!ht]
    \centering
    \begin{subfigure}[t]{0.45\textwidth}
        \centering
        \begin{minipage}[c]{0.33\linewidth}
            \includegraphics[width=\linewidth]{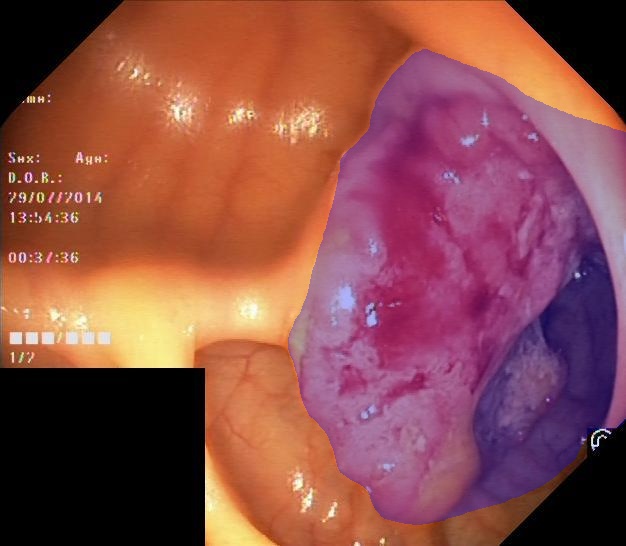}
        \end{minipage}
        \hfill
        \begin{minipage}[c]{0.64\linewidth}
            \scriptsize
            \raggedright
            \begin{tabularx}{\linewidth}{@{}lX@{}}
            \textbf{Pred:} & polyp measuring 11 to 20 millimeters \\
            \textbf{Exp:} & -- \\
            \end{tabularx}
        \end{minipage}
    \end{subfigure}
    \hfill
    \begin{subfigure}[t]{0.45\textwidth}
        \centering
        \begin{minipage}[c]{0.33\linewidth}
            \includegraphics[width=\linewidth]{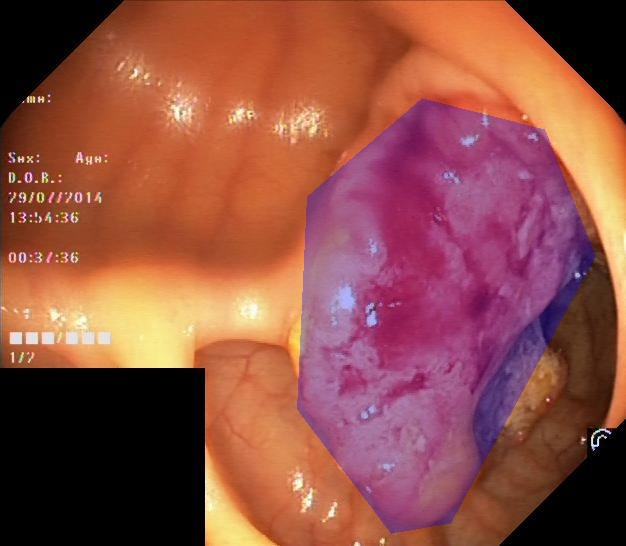}
        \end{minipage}
        \hfill
        \begin{minipage}[c]{0.64\linewidth}
            \scriptsize
            \raggedright
            \begin{tabularx}{\linewidth}{@{}lX@{}}
            \textbf{Pred:} & polyp measuring greater than 20 millimeters \\
            \textbf{Exp:} & The polyp measures greater than 20 millimeters in size. It appears as a large, rounded, and irregular shape with a mix of red, pink, and yellow colors. \\
            \end{tabularx}
        \end{minipage}
    \end{subfigure}

    \vspace{0.5em}
    \caption{
    Comparison of model responses before and after multi-task training.
    \textbf{Question:} What is the size of the polyp? 
    \textbf{Actual Answer:} polyp larger than 20 millimeters.
    }
    \label{fig:vqa_comparison}
\end{figure}

\begin{figure}[!ht]
    \centering
    \includegraphics[width=0.65\textwidth]{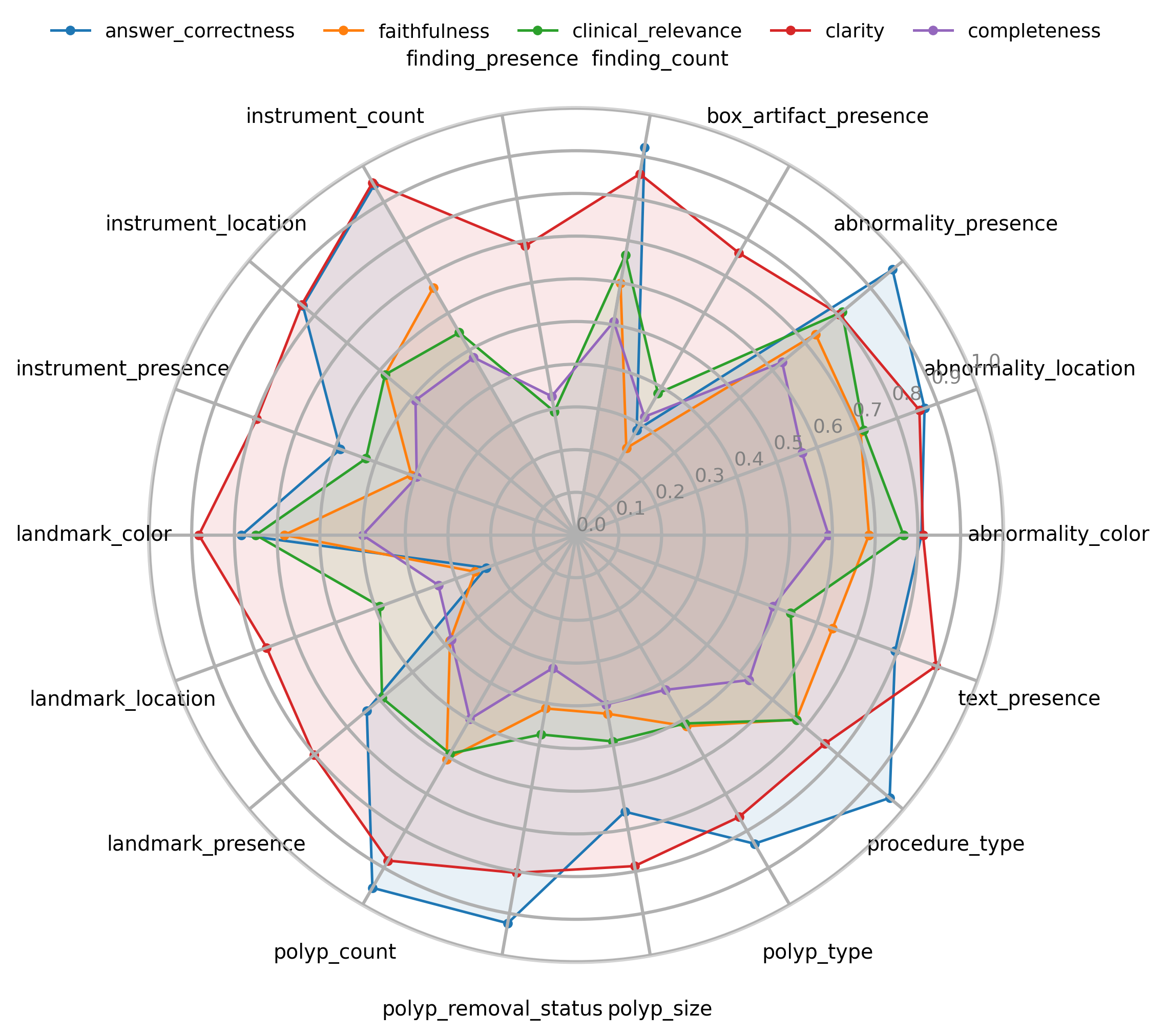}
    \caption{Sub-Task 2: Question-wise radar graph for each explainability metric on official results.}
    \label{fig:sub-task2}
\end{figure}

\section{Results and Evaluation}

Increasing LoRA parameters, as referenced in Table~\ref{tab:validation_scores}, reveals a critical trade-off in VQA-only training. While language scores improved, with BLEU increasing from $0.4348$ to $0.4623$, visual grounding severely degraded as Seg-IoU Instrument scores dropped from $0.4911$ to $0.2961$. This suggests the model overfits to linguistic cues. Conversely, our multi-task approach improved all metrics simultaneously; Seg-IoU Instrument scores rose from $0.7098$ to $0.7403$ and BLEU scores climbed from $0.4432$ to $0.4726$. This culminated in our best-performing model (FL2\_VQA\_MT\_128\_256), which demonstrated robust generalization to unseen data on the official private dataset with a BLEU of $0.4539$, ROUGE-L of $0.6531$, and BERTScore F1 of $0.9479$. This shows the segmentation task acts as an effective regularizer, preserving visual understanding while enhancing language quality. This is evident in Figure~\ref{fig:vqa_comparison}, where the multi-task model produces a more accurate polyp segmentation and correctly identifies it as “greater than 20 millimeters,” consistent with the ground truth.

As shown in the radar chart (Figure~\ref{fig:sub-task2}), our model's textual explanations for Sub-Task 2 are a mixed success. For structured tasks like presence detection and counting, the model achieves high \textbf{answer correctness}  and \textbf{clarity} . However, it shows significant weaknesses in \textbf{completeness} (0.3--0.7) and \textbf{faithfulness} (0.4--0.6). The model excels with binary and quantitative attributes but struggles with descriptive ones, particularly color identification, where all metrics fall below 0.6. These results highlight that although our training has improved correctness, further refinement is essential to ensure explanations are complete and clinically reliable.
\section{Conclusion and Future Work}
Our work advances grounded training for VLMs in GI endoscopy but revealed key challenges. Synthetic explaination data generation using the Gemma-27B model produced less diverse descriptions, hindering effective training. Furthermore, severe class imbalance hindered visual grounding, as rare landmarks were overshadowed by prevalent polyps, and the lack of negative examples in the augmented datasets may have also biased the model training. Future work could leverage the model's pseudo-masking to propose regions for refinement, using them as prompts for the Segment Anything Model (SAM) \cite{kirillov2023segment}, which can generate more precise masks that may be used to train an auxiliary segmentation head alongside VQA training. Additional improvements can be made by dataset augmentation for rare classes, incorporating negative examples, and grounding explanations by expert descriptions.

\section*{Declaration on Generative AI}
Generative AI tools were used for paraphrasing and improving grammatical flow, as well as assisting in diagram and table refinement. All content was carefully verified by the authors.

\clearpage

\def\bibfont{\small}
\bibliography{references}

\begin{document}
\copyrightyear{2025}
\copyrightclause{Copyright for this paper by its authors.
  Use permitted under Creative Commons License Attribution 4.0
  International (CC BY 4.0).}

\conference{MediaEval'25: Multimedia Evaluation Workshop,
  October 25--26, 2025, Dublin, Ireland and Online}

\title{Multi-Task Learning for Visually Grounded Reasoning in Gastrointestinal VQA} 

\author[1]{Itbaan Safwan}[%
  email=i.safwan.26197@khi.iba.edu.pk,
]
\author[1]{Muhammad Annas Shaikh}[%
  email=m.shaikh.26919@khi.iba.edu.pk,
]
\author[1]{Muhammad Haaris}[%
  email=m.haaris.27083@khi.iba.edu.pk,
]

\author[1]{Ramail Khan}[%
  email=r.khan.26924@khi.iba.edu.pk,
]

\author[1]{Muhammad Atif Tahir}[%
  email=atiftahir@iba.edu.pk,
]

\address[1]{School of Mathematics and Computer Science, Institute of Business Administration (IBA), Karachi, Pakistan}

\begin{abstract}
We present a multi-task framework for the MediaEval Medico 2025 challenge, leveraging a LoRA-tuned Florence-2 model for simultaneous visual question answering (VQA), explanation generation, and visual grounding. The proposed system integrates three curated datasets: (1) Kvasir-VQA-x1 for question-answer learning, (2) a synthetically enriched explanation dataset offering structured medical reasoning, and (3) text-to-region pairs linking visual features with segmentation masks. This multi-task setup enables the model to jointly learn visual grounding, reasoning, and interpretation, producing responses that are both accurate and interpretable. Extensive evaluation demonstrates that our approach substantially improves over single-task baselines in both answer accuracy and visual localization, highlighting the effectiveness of grounded multi-task learning for medical VQA applications. Our code: \href{https://github.com/itbaans/multi_task_training_gastrointestinal_vqa}{\texttt{GitHub Repository}}

\end{abstract}

\maketitle

\input{sections/3.TaskOverviewandDataset}

\input{sections/6.AblationStudies}
\input{sections/7.Discussion}
\section{Conclusion and Future Work}
Our work advances grounded training for VLMs in GI endoscopy but revealed key challenges. Synthetic explaination data generation using the Gemma-27B model produced less diverse descriptions, hindering effective training. Furthermore, severe class imbalance hindered visual grounding, as rare landmarks were overshadowed by prevalent polyps, and the lack of negative examples in the augmented datasets may have also biased the model training. Future work could leverage the model's pseudo-masking to propose regions for refinement, using them as prompts for the Segment Anything Model (SAM) \cite{kirillov2023segment}, which can generate more precise masks that may be used to train an auxiliary segmentation head alongside VQA training. Additional improvements can be made by dataset augmentation for rare classes, incorporating negative examples, and grounding explanations by expert descriptions.

\section*{Declaration on Generative AI}
Generative AI tools were used for paraphrasing and improving grammatical flow, as well as assisting in diagram and table refinement. All content was carefully verified by the authors.

\clearpage

\input{sections/Acknowledgments}

\def\bibfont{\small} 
\bibliography{references} 


\end{document}